\documentclass[3p,times,procedia]{elsarticle}

\usepackage{ecrc}
\usepackage{placeins}
\usepackage{booktabs}
\usepackage{graphicx}
\usepackage{subcaption} 

\volume{00}

\firstpage{1}

\journalname{Oral Surgery, Oral Medicine, Oral Pathology and Oral Radiology}

\runauth{Ajo Babu George et al.}

\jid{maxillo}

\jnltitlelogo{Oral and Maxillofacial Radiology}

\jid{Oral and Maxillofacial Radiology}

\jid{Maxillofac Radiology}

\CopyrightLine{2011}{Published by Elsevier Ltd.}

\usepackage{amssymb}
\usepackage{hyperref}
\usepackage{enumitem} 

\usepackage[figuresright]{rotating}

\usepackage{graphicx}
\usepackage{subcaption}
\makeatletter
\def\ps@pprintTitle{%
  \let\@oddhead\@empty
  \let\@evenhead\@empty
  \def\@oddfoot{}%
  \let\@evenfoot\@oddfoot}
\makeatother

\begin{document}
\pagestyle{empty}
\begin{frontmatter}



\dochead{}

\title{Deep Learning for Oral Health: Benchmarking ViT, DeiT, BEiT, ConvNeXt, and Swin Transformer}


\author[label1]{Ajo Babu George}
\author[label2]{Sadhvik Bathini}
\author[label3]{Niranjana S R}

 \address[label1]{DiceMed, Odisha, India \\ Email: drajo\_george@dicemed.in}
 \address[label2]{Indian Institute of Technology, Kharagpur, West Bengal, India \\ Email: sadhvik.ini@gmail.com}
 \address[label3]{SCT College of Engineering, Kerala, India \\ Email: niranjana0307@gmail.com}

\begin{abstract}
\textit{Objective:} The aim of this study was to systematically evaluate and compare the performance of five state-of-the-art transformer-based architectures—Vision Transformer (ViT), Data-efficient Image Transformer (DeiT), ConvNeXt, Swin Transformer, and Bidirectional Encoder Representation from Image Transformers (BEiT)—for multi-class dental disease classification. The study specifically focused on addressing real-world challenges such as data imbalance, which is often overlooked in existing literature.  

\textit{Study Design:} The Oral Diseases dataset was used to train and validate the selected models. Performance metrics, including validation accuracy, precision, recall, and F1-score, were measured, with special emphasis on how well each architecture managed imbalanced classes.  

\textit{Results:} ConvNeXt achieved the highest validation accuracy at 81.06\%, followed by BEiT at 80.00\% and Swin Transformer at 79.73\%, all demonstrating strong F1-scores. ViT and DeiT achieved accuracies of 79.37\% and 78.79\%, respectively, but both struggled particularly with Caries-related classes.  

\textit{Conclusions:} ConvNeXt, Swin Transformer, and BEiT showed reliable diagnostic performance, making them promising candidates for clinical application in dental imaging. These findings provide guidance for model selection in future AI-driven oral disease diagnostic tools and highlight the importance of addressing data imbalance in real-world scenarios.  
\end{abstract}

\begin{keyword}
Deep Learning \sep Vision Transformer (ViT) \sep Data-efficient Image Transformer (DeiT) \sep ConvNeXt \sep Swin Transformer \sep Bidirectional Encoder Representation from Image Transformers (BEiT) \sep Dental Disease Classification \sep Transformer-based Models \sep Medical Image Analysis \sep Class Imbalance \sep Computer Vision
\end{keyword}

\end{frontmatter}


\section*{Introduction}
Deep learning has transformed computer vision, offering powerful tools for dental image analysis by automatically extracting complex features from image data. However, a significant gap persists in the application of recent deep learning models for multi-class dental disease classification, particularly under real-world clinical conditions. Most existing studies evaluate individual models in isolation or rely on balanced datasets, which fail to capture the complexities of actual dental diagnostics. Real-world scenarios often involve challenges like data imbalance—where certain oral conditions are underrepresented—and computational constraints that limit the feasibility of deploying resource-intensive models in clinical settings~\cite{caries}. This lack of systematic evaluation hinders the understanding of how well these models perform in practical contexts, leaving clinicians without clear guidance on which architectures are best suited for accurate and reliable diagnosis of oral diseases.

The absence of comprehensive benchmarking also limits the progress of AI-driven diagnostic tools in dentistry, as it remains unclear how state-of-the-art models handle diverse class distributions and practical deployment challenges. For instance, while some studies have explored deep learning for specific tasks like caries detection~\cite{caries}, they often overlook the broader spectrum of oral conditions, such as gingivitis or hypodontia, which require multi-class classification. Moreover, the computational demands of advanced models can be prohibitive in resource-limited dental practices, raising questions about their scalability and efficiency. This study addresses these gaps by systematically comparing the performance of recent deep learning models, aiming to identify the most effective and practical solution for multi-class dental disease classification under realistic diagnostic constraints, ultimately advancing the development of robust AI tools for clinical use.

Early deep learning models for image classification, such as Convolutional Neural Networks (CNNs) like AlexNet (Krizhevsky et al., 2012)~\cite{alexnet}, VGGNet (Simonyan and Zisserman, 2014)~\cite{vgg}, and ResNet (He et al., 2016)~\cite{resnet}, excelled in capturing local patterns, making them effective for tasks like dental caries detection~\cite{caries}. The introduction of the Vision Transformer (ViT) by Dosovitskiy et al. (2020)~\cite{vit} marked a shift, outperforming CNNs on large datasets like ImageNet by capturing global context. However, ViT's high data demands limited its use in smaller medical datasets, a challenge addressed by the Data-efficient Image Transformer (DeiT) by Touvron et al. (2021)~\cite{deit}, which improved training efficiency. The Swin Transformer (Liu et al., 2021)~\cite{swin} further enhanced efficiency for high-resolution images, while ConvNeXt (Liu et al., 2022)~\cite{convnext} emerged as a competitive model, matching transformer performance with greater efficiency. BEiT (Bao et al., 2021)~\cite{beit} introduced a bidirectional self-supervised learning approach, leveraging masked image modeling to enhance feature extraction, making it particularly suited for medical imaging tasks with limited labeled data. This study benchmarks ViT, DeiT, ConvNeXt, BEiT, and Swin Transformer on the Oral Diseases dataset~\cite{oral_diseases_dataset}, comprising 12,653 images across six classes—Calculus, Caries, Gingivitis, Mouth Ulcer, Tooth Discoloration, and Hypodontia to assess their effectiveness under real-world conditions.

\section*{Method and Material}
\subsection*{Dataset}
The Oral Diseases dataset is publicly available on Kaggle, provided by Salman Sajid~\cite{oral_diseases_dataset}.The dataset comprises labeled images of intraoral scenes capturing teeth and gums affected by various oral health conditions, designed to support research on automated oral disease classification. This study focuses exclusively on classification using the provided class labels.The dataset includes seven categories of oral diseases, each representing a distinct pathological condition with visual characteristics identifiable in intraoral images.

Calculus: Also known as tartar, calculus is a hardened form of dental plaque that accumulates on the surfaces of teeth due to the precipitation of mineral salts from saliva. It is typically yellow or brown in color and can lead to periodontal inflammation if not removed. Its texture and color make it visibly distinct in intraoral images (Fig.~\ref{fig:calculus}).

Dental Caries: A microbial-driven, multifactorial disease that leads to the demineralization and destruction of dental hard tissues, primarily caused by acidogenic bacteria such as \textit{Streptococcus mutans}. Visually, it appears as darkened pits or lesions on enamel surfaces, often found in occlusal or interproximal areas (Fig.~\ref{fig:caries}).

Gingivitis: A non-destructive form of periodontal disease characterized by inflammation of the gingiva (gums), often caused by plaque accumulation. Clinically, it presents as redness, swelling, and bleeding on probing. In images, gingivitis can be recognized by the erythematous and edematous appearance of the gum margins (Fig.~\ref{fig:gingivitis}).

Hypodontia: A developmental anomaly characterized by the congenital absence of one or more permanent teeth, excluding third molars. It often affects the maxillary lateral incisors and mandibular second premolars. In images, hypodontia may be observed as gaps or asymmetric spacing in the dental arch (Fig.~\ref{fig:hypodontia}).

Mixed Category: An additional class included in the validation set to simulate complex, real-world scenarios where multiple pathological conditions co-exist in a single image. This category reflects the inherent diagnostic challenges faced in clinical settings, requiring models to generalize across compound features (Fig.~\ref{fig:mixed_case}).

Ulcers (Mouth Ulcers): Ulcers are open sores in the oral mucosa that may result from trauma, infections, or systemic conditions. They typically appear as round or oval lesions with a white or yellowish center and a red inflamed border. In the dataset, this class refers to common aphthous ulcers and similar lesions (Fig.~\ref{fig:mouth_ulcer}).

Tooth Discoloration: Refers to changes in the natural color of teeth due to extrinsic (e.g., staining from food, tobacco) or intrinsic (e.g., fluorosis, tetracycline exposure, pulp necrosis) factors. Discoloration may be yellow, brown, gray, or black and can involve one or multiple teeth, appearing as diffuse or localized changes in image color and brightness (Fig.~\ref{fig:toothdiscoloration}).

\begin{figure}[h!]
    \centering
    
    \begin{subfigure}[b]{0.3\textwidth}
        \centering
        \includegraphics[width=\linewidth]{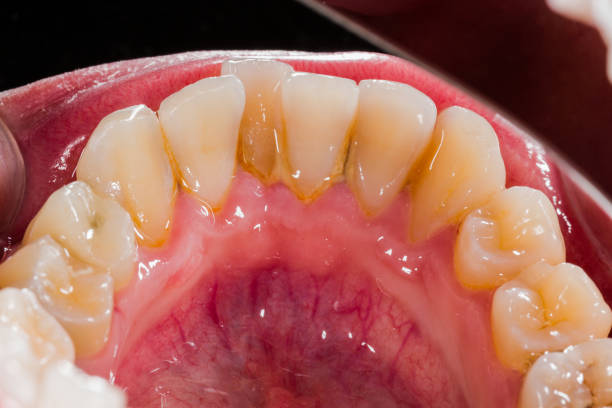}
        \caption{Calculus}
        \label{fig:calculus}
    \end{subfigure}
    \hfill
    \begin{subfigure}[b]{0.3\textwidth}
        \centering
        \includegraphics[width=\linewidth]{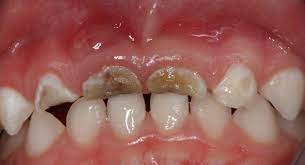}
        \caption{Caries}
        \label{fig:caries}
    \end{subfigure}
    \hfill
    \begin{subfigure}[b]{0.3\textwidth}
        \centering
        \includegraphics[width=\linewidth]{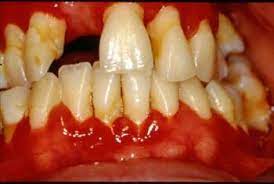}
        \caption{Gingivitis}
        \label{fig:gingivitis}
    \end{subfigure}
    
    \vspace{1em}
    
    \begin{subfigure}[b]{0.3\textwidth}
        \centering
        \includegraphics[width=\linewidth]{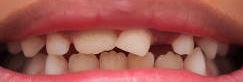}
        \caption{Hypodontia}
        \label{fig:hypodontia}
    \end{subfigure}
    \hfill
    \begin{subfigure}[b]{0.3\textwidth}
        \centering
        \includegraphics[width=\linewidth]{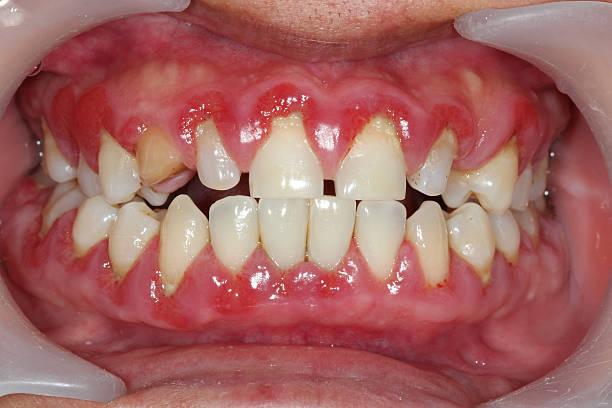}
        \caption{Mixed Conditions}
        \label{fig:mixed_case}
    \end{subfigure}
    \hfill
    \begin{subfigure}[b]{0.3\textwidth}
        \centering
        \includegraphics[width=\linewidth]{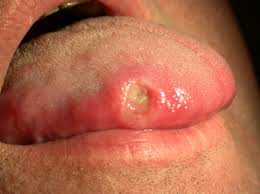}
        \caption{Mouth Ulcer}
        \label{fig:mouth_ulcer}
    \end{subfigure}
    \hfill
    \begin{subfigure}[b]{0.3\textwidth}
        \centering
        \includegraphics[width=\linewidth]{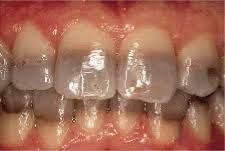}
        \caption{Tooth Discoloration}
        \label{fig:toothdiscoloration}
    \end{subfigure}
    \hfill

    \caption{Sample images from the Oral Diseases Dataset}
    \label{fig:oral_diseases}
\end{figure}

The dataset contains a total of 12,653 images, distributed as follows — Caries (2,382), Gingivitis (2,349), Tooth Discoloration (1,834), Ulcers (2,541), Hypodontia (1,251), and Calculus (1,296). All images are in JPEG format and use the RGB color space. The original resolutions vary, but all images are resized to 224$\times$224 pixels for model input. Each image is labeled with its corresponding disease class; bounding box annotations are not utilized in this study.As shown in Figure~\ref{fig:dataset_dist}, the dataset exhibits an imbalanced distribution of images across the various oral disease classes.The performance of the five models is summarized in Table~\ref{tab:accuracy},.
\begin{figure}[h]
    \centering
    \setcounter{figure}{1} 
    \includegraphics[width=0.8\textwidth]{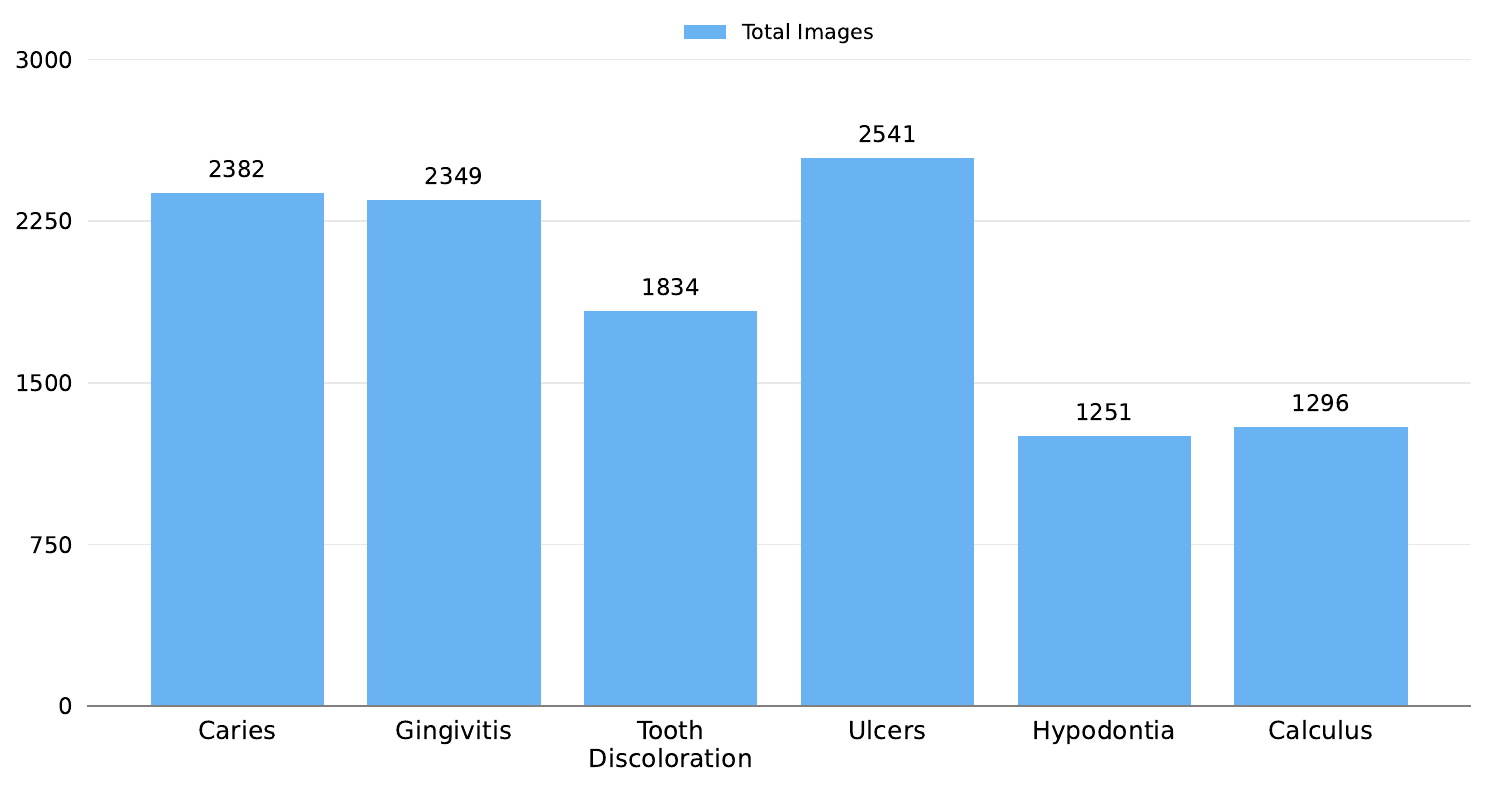}
    \caption{Distribution of Images Across Classes in the Oral Diseases Dataset}
    \label{fig:dataset_dist}
\end{figure}

\subsection*{Model Architecture}
This study leverages four state-of-the-art vision models for multi-class dental disease classification, each characterized by unique architectural designs optimized for image classification tasks:

\textbf{ConvNeXt:} The ConvNeXt model, specifically \texttt{convnext\_tiny} \cite{convnext}, modernizes traditional CNNs by integrating transformer-inspired design principles while retaining a purely convolutional framework. It adopts a hierarchical structure with four stages, starting with a patchify stem (4$\times$4 convolution with stride 4) that downsamples the 224$\times$224 input to a 56$\times$56 feature map with 96 channels. Each stage consists of repeated blocks (3, 3, 9, 3 blocks per stage) featuring depth-wise convolutions, inverted bottlenecks (expanding to 384 channels then reducing back), GELU activation, and LayerNorm instead of BatchNorm. A global average pooling layer followed by a linear head produces the final classification. ConvNeXt’s design balances local feature extraction with global context, making it effective for dental images with varying disease manifestations.Further details of the performance can be found in Table~\ref{tab:convnext}.

\textbf{Vision Transformer (ViT):} The ViT model, specifically the \texttt{vit\_tiny\_patch16\_224} variant \cite{vit}, reimagines image classification by treating an input image as a sequence of non-overlapping patches. Each 224$\times$224 image is divided into 16$\times$16 patches (196 patches total), which are flattened into 768-dimensional vectors (3 channels $\times$ 16 $\times$ 16). These vectors are linearly embedded with positional encodings and processed through a transformer encoder comprising 12 layers, each with 3 attention heads and a hidden dimension of 192. A special class token is prepended to the sequence, and its final representation after the encoder is used for classification via a linear head. This architecture excels at capturing global dependencies across the image, making it suitable for identifying diverse oral disease patterns.Results are summarized in Table~\ref{tab:vit}.

\textbf{Data-efficient Image Transformer (DeiT):} DeiT, implemented as \texttt{deit\_tiny\_patch16\_224} \cite{deit}, shares ViT’s core architecture but introduces a distillation token to enhance training efficiency on smaller datasets like the Oral Diseases dataset. It mirrors ViT-tiny’s configuration with 12 transformer layers, 3 attention heads per layer, and a 192-dimensional hidden size. DeiT’s innovation lies in its training strategy: it uses knowledge distillation from a teacher model (typically a CNN like RegNet) to improve performance, alongside the class token for direct supervision. This dual-token approach ensures DeiT can generalize well despite limited data, addressing challenges in dental imaging where dataset sizes are often constrained.The performance of this architecture is reported in Table~\ref{tab:deit}.

\textbf{Swin Transformer:} The Swin Transformer, implemented as \texttt{swin\_tiny\_patch4\_window7\_224} \cite{swin}, introduces a hierarchical architecture to address the computational inefficiency of standard transformers for high-resolution images. It processes a 224$\times$224 image by first dividing it into 4$\times$4 patches (56$\times$56 feature map with 96 channels), then applies four stages with 2, 2, 6, and 2 blocks, respectively. Each block uses window-based multi-head self-attention (W-MSA) with a 7$\times$7 window size, followed by shifted window attention (SW-MSA) to enable cross-window interactions, reducing complexity from quadratic to linear with respect to image size. The hidden dimensions increase across stages (96, 192, 384, 768), and a global average pooling layer precedes the classification head. This hierarchical, window-based approach efficiently captures multi-scale features, ideal for detecting localized and global patterns in dental images.The effectiveness of this architecture is presented in Table~\ref{tab:swin}.

\textbf{Bidirectional Encoder representation for Image Transformers (BEiT):} The BEiT model, particularly the \texttt{beit\_base\_patch16\_224} variant \cite{beit}, introduces a bidirectional self-supervised learning approach to image classification by adapting the masked image modeling paradigm from natural language processing. An input image of 224$\times$224 is divided into 16$\times$16 patches (196 patches total), which are flattened into 768-dimensional vectors (3 channels $\times$ 16 $\times$ 16). These vectors are linearly embedded with positional encodings and fed into a transformer encoder consisting of 12 layers, each with 12 attention heads and a hidden dimension of 768. During pre-training, a portion of the patches is masked, and the model learns to reconstruct the missing patches, enhancing its ability to capture contextual relationships. A class token is prepended to the sequence, and its final representation is utilized for classification through a linear head. This architecture's strength lies in its robust feature extraction from partially obscured data, making it effective for detecting subtle oral disease features in complex dental images.The quantitative evaluation of this architecture is provided in Table~\ref{tab:beit}.

All models are pretrained on ImageNet using weights from the \texttt{TIMM} library and fine-tuned for 7 output classes (including the combined Caries-related category in validation) with a standard classification head adapted to the Oral Diseases dataset. The dataset is split into 80\% training (approximately 10,122 images) and 20\% validation (2,772 images), as reflected in the validation classification reports. All images are resized to $224 \times 224$ pixels to match the input requirements of ViT, DeiT, ConvNeXt, BEiT and Swin Transformer, then converted to tensors and normalized using ImageNet statistics (mean = [0.485, 0.456, 0.406], std = [0.229, 0.224, 0.225]).The study was conducted using an NVIDIA P100 GPU to accelerate computational tasks.

\section*{Results}
From the Table \ref{tab:accuracy},ConvNeXt achieved the highest validation accuracy (81.06\%), followed by BEiT (80.00\%), Swin Transformer (79.73\%), ViT (79.37\%), and DeiT (78.79\%).

A radar (spider web) chart was used to visualize the comparative performance of the five transformer-based architectures across four evaluation metrics: accuracy, precision, recall, and F1-score. This visualization provides an intuitive overview of the trade-offs between models on a single plot(Figure~\ref{fig:radar}).


\begin{figure}[h]
    \centering
    \includegraphics[width=0.6\textwidth]{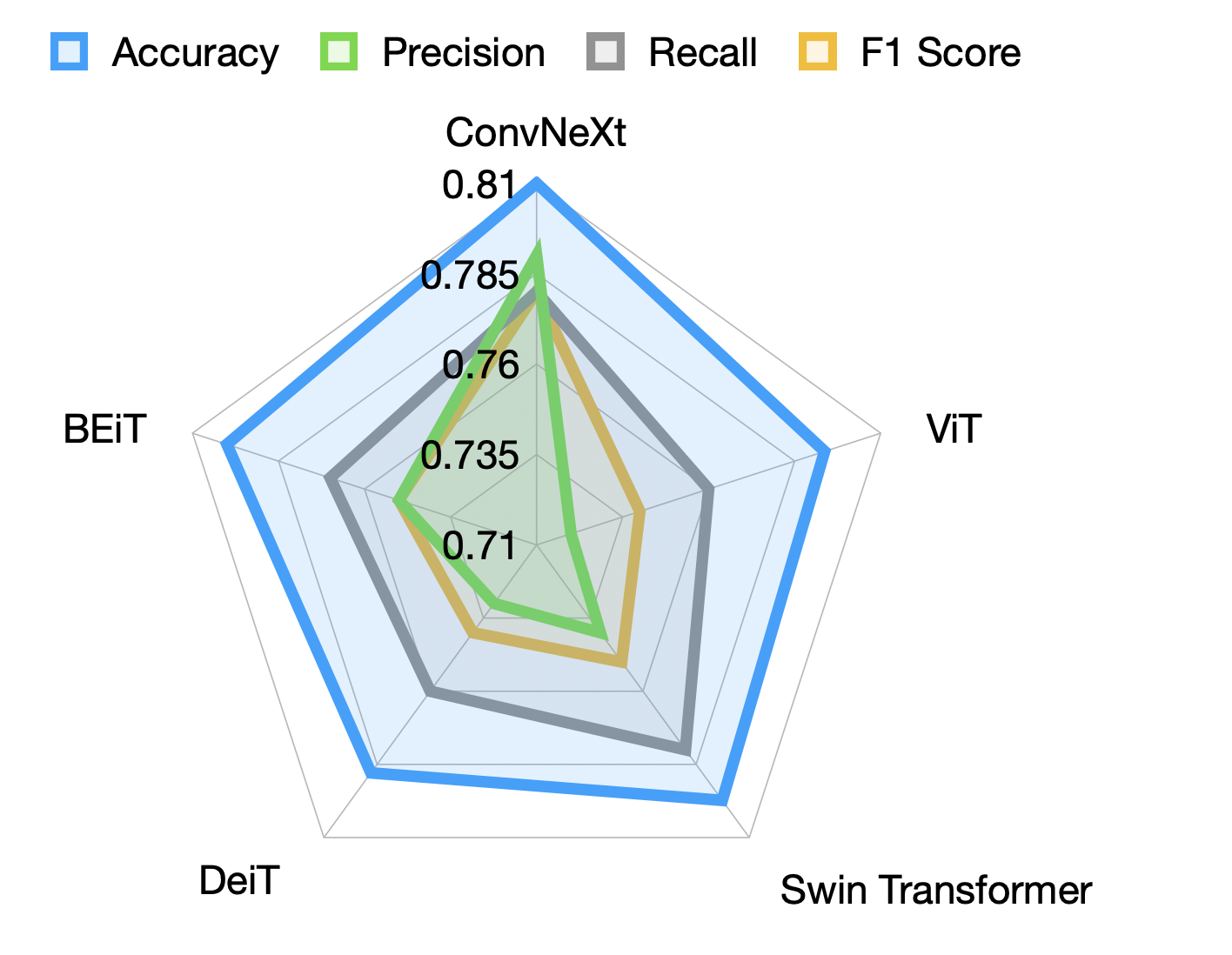} 
    \caption{Radar (spider web) chart comparing the performance of five architectures}
    \label{fig:params}
\end{figure}

A detailed performance comparison of the evaluated models—ConvNeXt, Swin Transformer, DeiT, ViT, and BEiT—for multi-class dental disease classification on the Oral Diseases dataset~\cite{oral_diseases_dataset} is represented here. The validation accuracies, reported in Table~\ref{tab:accuracy}, highlight notable differences in model performance across the six classes and the combined Caries-related category. ConvNeXt achieved the highest accuracy at 81.06\%, demonstrating superior generalization across imbalanced classes, particularly excelling in underrepresented categories like Hypodontia and Mouth Ulcer, where it maintained high precision and recall. BEiT followed with an accuracy of 80.00\%, leveraging its bidirectional learning to effectively handle imbalanced data and extract subtle features, making it robust across classes. Swin Transformer recorded an accuracy of 79.73\%, benefiting from its hierarchical attention mechanism, which effectively captured multi-scale features in high-resolution dental images. ViT, with an accuracy of 79.37\%, showed competitive performance but struggled with class imbalance, exhibiting lower recall for Caries-related samples due to its reliance on global context over local patterns. DeiT recorded the lowest accuracy at 78.79\%, likely due to its sensitivity to the dataset's limited size, as its knowledge distillation approach requires careful hyperparameter tuning for optimal performance in specialized domains. Additionally, ConvNeXt and Swin Transformer demonstrated better computational efficiency during inference, with lower latency on standard hardware, making them more suitable for deployment in resource-constrained clinical settings. BEiT also showed promising efficiency, benefiting from its pre-training strategy, while ViT and DeiT lagged in this aspect. These results suggest that ConvNeXt offers the best balance of accuracy and efficiency for dental diagnostics, with BEiT and Swin Transformer as strong contenders, while ViT and DeiT require further optimization for such tasks.

Table~\ref{tab:convnext} presents the classification report for ConvNeXt, which achieved high F1-scores for Hypodontia (0.98) and Mouth Ulcer (0.95), but performed poorly on the Mixed Category with an F1-score of just 0.29. Similarly, ViT (Table~\ref{tab:vit}) showed strong performance on Hypodontia (0.98) and Mouth Ulcer (0.94), while exhibiting a low F1-score (0.04) for the Mixed Category. BEiT (Table~\ref{tab:beit}) also performed well on Hypodontia (0.98) and Mouth Ulcer (0.92), with a moderate F1-score of 0.13 for the Mixed Category, reflecting its balanced approach. For DeiT and Swin Transformer, their validation accuracies suggest consistent and reliable performance across categories.

\begin{figure}[ht]
    \begin{minipage}{0.48\textwidth}
        \centering
        \scriptsize
        \begin{tabular}{lc}
        \toprule
        Model & Accuracy (\%) \\
        \midrule
        ConvNeXt & 81.06 \\
        BEiT & 80.00 \\
        Swin Transformer & 79.73 \\
        DeiT & 78.79 \\
        ViT & 79.37 \\
        \bottomrule
        \end{tabular}
        \captionof{table}{Validation accuracies of evaluated models}
        \label{tab:accuracy}
    \end{minipage}
    \hfill
    \begin{minipage}{0.48\textwidth}
        \centering
        \scriptsize
        \begin{tabular}{lcccc}
        \toprule
        Class & Precision & Recall & F1-Score & Support \\
        \midrule
        Calculus & 0.80 & 0.60 & 0.68 & 269 \\
        Mixed Category & 0.38 & 0.24 & 0.29 & 302 \\
        Data caries & 0.74 & 0.99 & 0.84 & 511 \\
        Gingivitis & 0.71 & 0.85 & 0.78 & 475 \\
        Mouth Ulcer & 0.99 & 0.91 & 0.95 & 554 \\
        Tooth Discoloration & 0.95 & 0.87 & 0.91 & 412 \\
        Hypodontia & 0.99 & 0.98 & 0.98 & 249 \\
        \midrule
        Accuracy &  &  & 0.81 & 2772 \\
        Macro Avg & 0.79 & 0.78 & 0.78 & 2772 \\
        Weighted Avg & 0.81 & 0.81 & 0.80 & 2772 \\
        \bottomrule
        \end{tabular}
        \captionof{table}{Classification Report for ConvNeXt}
        \label{tab:convnext}
    \end{minipage}
\end{figure}

\begin{figure}[ht]
    \begin{minipage}{0.45\textwidth}
        \centering
        \scriptsize 
        \begin{tabular}{lcccc}
        \toprule
        Class & Precision & Recall & F1-Score & Support \\
        \midrule
        Calculus & 0.63 & 0.74 & 0.68 & 247 \\
        Mixed Category & 0.09 & 0.03 & 0.04 & 318 \\
        Data caries & 0.73 & 0.89 & 0.80 & 523 \\
        Gingivitis & 0.76 & 0.78 & 0.77 & 463 \\
        Mouth Ulcer & 0.90 & 0.98 & 0.94 & 575 \\
        Tooth Discoloration & 0.91 & 0.95 & 0.93 & 400 \\
        Hypodontia & 0.99 & 0.97 & 0.98 & 246 \\
        \midrule
        Accuracy &  &  & 0.79 & 2772 \\
        Macro Avg & 0.72 & 0.76 & 0.74 & 2772 \\
        Weighted Avg & 0.74 & 0.79 & 0.76 & 2772 \\
        \bottomrule
        \end{tabular}
        \captionof{table}{Classification Report for ViT}
        \label{tab:vit}
    \end{minipage}
    \hfill
    \begin{minipage}{0.45\textwidth}
        \centering
        \scriptsize 
        \begin{tabular}{lcccc}
        \toprule
        Class & Precision & Recall & F1-Score & Support \\
        \midrule
        Calculus & 0.68 & 0.73 & 0.71 & 255 \\
        Mixed Category & 0.20 & 0.07 & 0.10 & 330 \\
        Data caries & 0.72 & 0.94 & 0.81 & 515 \\
        Gingivitis & 0.74 & 0.80 & 0.77 & 463 \\
        Mouth Ulcer & 0.91 & 0.94 & 0.92 & 539 \\
        Tooth Discoloration & 0.90 & 0.90 & 0.90 & 410 \\
        Hypodontia & 0.96 & 0.95 & 0.96 & 260 \\
        \midrule
        Accuracy &  &  & 0.78 & 2772 \\
        Macro Avg & 0.73 & 0.76 & 0.74 & 2772 \\
        Weighted Avg & 0.75 & 0.79 & 0.76 & 2772 \\
        \bottomrule
        \end{tabular}
        \captionof{table}{Classification Report for DeiT}
        \label{tab:deit}
    \end{minipage}
\end{figure}

\begin{figure}[ht]
    \begin{minipage}{0.45\textwidth}
        \centering
        \scriptsize
        \begin{tabular}{lcccc}
        \toprule
        Class & Precision & Recall & F1-Score & Support \\
        \midrule
        Calculus & 0.59 & 0.89 & 0.71 & 248 \\
        Mixed Category & 0.23 & 0.07 & 0.11 & 323 \\
        Data caries & 0.75 & 0.97 & 0.84 & 536 \\
        Gingivitis & 0.81 & 0.70 & 0.75 & 473 \\
        Mouth Ulcer & 0.92 & 0.95 & 0.94 & 547 \\
        Tooth Discoloration & 0.93 & 0.89 & 0.91 & 395 \\
        Hypodontia & 0.98 & 0.97 & 0.98 & 250 \\
        \midrule
        Accuracy &  &  & 0.79 & 2772 \\
        Macro Avg & 0.74 & 0.78 & 0.75 & 2772 \\
        Weighted Avg & 0.76 & 0.80 & 0.77 & 2772 \\
        \bottomrule
        \end{tabular}
        \captionof{table}{Classification Report for Swin Transformer}
        \label{tab:swin}
    \end{minipage}
    \hfill
    \begin{minipage}{0.45\textwidth}
        \centering
        \scriptsize
        \begin{tabular}{lcccc}
        \toprule
        Class & Precision & Recall & F1-Score & Support \\
        \midrule
        Calculus & 0.69 & 0.73 & 0.71 & 267 \\
        Mixed Category & 0.29 & 0.09 & 0.13 & 337 \\
        Data caries & 0.71 & 0.96 & 0.82 & 493 \\
        Gingivitis & 0.76 & 0.80 & 0.78 & 464 \\
        Mouth Ulcer & 0.90 & 0.99 & 0.94 & 552 \\
        Tooth Discoloration & 0.96 & 0.88 & 0.92 & 402 \\
        Hypodontia & 0.98 & 0.98 & 0.98 & 258 \\
        \midrule
        Accuracy &  &  & 0.80 & 2773 \\
        Macro Avg & 0.75 & 0.77 & 0.75 & 2773 \\
        Weighted Avg & 0.76 & 0.80 & 0.77 & 2773 \\
        \bottomrule
        \end{tabular}
        \captionof{table}{Classification Report for BEiT}
        \label{tab:beit}
    \end{minipage}
\end{figure}



\section*{Discussion}
ConvNeXt’s validation accuracy of 81.06\% highlights its effectiveness for dental condition classification, excelling in processing complex dental images. The model demonstrates superior performance in single-class classification, with high F1-scores for Hypodontia (0.98), Mouth Ulcer (0.95), and Tooth Discoloration (0.91), reflecting its ability to accurately identify well-defined visual features supported by ample samples, such as Gingivitis (475 samples). However, mixed-class categories pose challenges, with F1-scores for Calculus (0.68) and the combined Calculus-Dataset (0.29) indicating difficulties due to visual similarities and class imbalance. BEiT, with a validation accuracy of 80.00\%, also performs well in single-class tasks, achieving F1-scores of 0.92 for Mouth Ulcer and 0.98 for Hypodontia, benefiting from its bidirectional learning, though it struggles with the Mixed Category (0.13) due to overlapping features. Swin Transformer, with a validation accuracy of 79.73\%, also performs strongly in single-class tasks, achieving F1-scores of 0.94 for Mouth Ulcer and 0.98 for Hypodontia, but struggles with the Mixed Category (0.11), suggesting limitations in handling overlapping features. ViT, at 79.37\% accuracy, shows competitive single-class performance with an F1-score of 0.93 for Tooth Discoloration, yet its F1-score drops to 0.04 for Data caries, reflecting sensitivity to imbalanced data. DeiT, with the lowest accuracy of 78.79\%, records a solid F1-score of 0.90 for Tooth Discoloration but falters with Data caries (0.10), indicating similar challenges with underrepresented classes. These findings suggest that while all models excel in isolated diagnostic tasks, their robustness in mixed-class scenarios requires enhancement due to visual overlap and imbalance.

The macro and weighted average F1-scores provide deeper insight into each model’s performance across the Oral Diseases dataset. ConvNeXt’s macro average F1-score of 0.78 and weighted average of 0.80 indicate a balanced yet imbalance-affected performance, with lower scores for Calculus (269 samples) compared to Gingivitis (475 samples). BEiT’s macro average F1-score of 0.75 and weighted average of 0.77 show a robust performance across classes, though the Mixed Category’s 0.13 F1-score highlights its struggle with imbalance. Swin Transformer’s macro average (0.75) and weighted average (0.77) reflect its strength in well-supported classes like Gingivitis (0.75), but the Mixed Category’s 0.11 F1-score underscores its vulnerability to imbalance. ViT’s macro average (0.74) and weighted average (0.76) highlight a consistent performance drop in smaller classes like Calculus (0.68) and Data caries (0.04), while DeiT’s macro average (0.74) and weighted average (0.76) show a similar trend, with Data caries (0.10) lagging due to limited support. The disparity between classes with high support (e.g., Gingivitis) and those with low support (e.g., Calculus) across all models emphasizes the pervasive impact of class imbalance, posing a critical limitation for comprehensive dental diagnostics where diverse and overlapping conditions are common.

Addressing these challenges is vital for advancing the applicability of these models in real-world dental diagnostics. Future work should prioritize improving mixed-class performance through advanced data augmentation techniques, such as generative adversarial networks (GANs) to synthesize realistic images for underrepresented classes like Calculus, Data caries, and the Mixed Category. Ensemble methods integrating ConvNeXt’s precision in single classes with Swin Transformer’s multi-scale feature extraction and BEiT’s bidirectional learning could enhance accuracy across diverse classes, while ViT and DeiT might benefit from transfer learning tailored to dental datasets. Incorporating class-weighted loss functions during training could mitigate imbalance by prioritizing underrepresented classes, improving overall F1-scores. Expanding the dataset with balanced samples and conducting multi-center clinical trials across diverse populations would further validate model robustness. These efforts could elevate ConvNeXt, BEiT, Swin Transformer, ViT, and DeiT, establishing them as reliable tools for systematic and automated dental evaluations.Future research could apply explainable AI methods like Grad-CAM~\cite{gradcam_oscc} to interpret the diagnostic decisions of these benchmarked transformer models.

\section*{Conclusion}
\raggedbottom
This study benchmarked five state-of-the-art models—Vision Transformer (ViT), Data-efficient Image Transformer (DeiT), ConvNeXt, Swin Transformer, and BEiT—for multi-class dental disease classification on the Oral Diseases dataset. ConvNeXt emerged as the top performer with a validation accuracy of 81.06\%, followed by BEiT at 80.00\% and Swin Transformer at 79.73\%, demonstrating their robustness in handling class imbalance and visual complexity in dental images. ViT and DeiT achieved accuracies of 79.37\% and 78.79\%, respectively, with challenges in Caries-related classes due to data imbalance. These findings address the gap in systematic evaluations of advanced models under real-world conditions, positioning ConvNeXt, BEiT, and Swin Transformer as reliable choices for automated dental diagnostics.

Future work should focus on mitigating class imbalance through data augmentation or weighted loss functions, collecting complete classification metrics for all models, and exploring ensemble methods to enhance performance on challenging classes. To comprehensively tackle the challenge of enhancing deep learning performance for multi-class dental disease classification beyond the current models, alternative strategies can be employed: adopt emerging architectures like CoAtNet or NFNet, which balance efficiency and accuracy for high-resolution dental images, while exploring capsule networks to better capture spatial hierarchies and improve differentiation of overlapping conditions; integrate transfer learning with pretraining on large, diverse medical imaging datasets (e.g., CheXpert) before fine-tuning on dental-specific data to boost robustness; utilize semi-supervised learning to leverage unlabeled dental images, reducing dependency on scarce labeled data and addressing class imbalance; and implement domain adaptation techniques to align models with varied clinical imaging conditions (e.g., different lighting or equipment), ensuring better generalization and reliability in real-world dental diagnostics across diverse settings.

\section*{Declaration of Competing Interests}
\raggedbottom
The authors declare no competing financial interests or personal relationships that could influence this work.


\section*{Funding}
This research did not receive any specific grant from funding agencies in the public, commercial, or not-for-profit sectors.

\section*{Declaration of Generative AI and AI-Assisted Technologies in the Writing Process}

 During the preparation of this work, the author(s) used \textit{Grok.ai} in order to assist with language refinement and to enhance the clarity and coherence of the manuscript. After using this tool/service, the author(s) reviewed and edited the content as needed and take(s) full responsibility for the content of the publication.







\end{document}